\documentclass[a4paper,10pt]{article}
\pdfoutput=1
\usepackage[margin=1in]{geometry}
\usepackage[T1]{fontenc}
\usepackage[utf8]{inputenc}
\usepackage{authblk}
\usepackage{lmodern}
\usepackage{soul}
\usepackage[normalem]{ulem}
\usepackage[final]{changes}
\usepackage{setspace}
\usepackage{cancel}
\usepackage{placeins}
\usepackage{flafter} 
\usepackage{graphicx}
\usepackage{subcaption} 
\usepackage{tabularx} 
\usepackage{array}
\usepackage{multirow}
\usepackage{url}

\usepackage{amsmath}  
\usepackage{amssymb}
\usepackage{graphicx} 
\usepackage[noend,boxruled]{algorithm2e}
\newcommand{\CN}{\(C/N_{0}\)}
\DeclareUnicodeCharacter{2212}{-}
\usepackage[backend=biber,style=authoryear,citestyle=apa,sorting=nyt,sortcites=true
]{biblatex}
\makeatletter
\AtEveryCitekey{\ifnameundef{shortauthor}{}{\def\cbx@apa@ifnamesaved{\@firstoftwo}}}
\makeatother
\usepackage{subfiles} 
\begin{document}

\title{3D map creation using crowdsourced
GNSS data}
\author[1]{Terence Lines}
\author[1]{Ana Basiri}
\affil[1]{School of Geographical \& Earth Sciences, University of Glasgow, United Kingdom}
\date{30 May 2020}

\maketitle

\begin{abstract}
3D maps are increasingly useful for many applications such as drone navigation, emergency services, and urban planning. However, creating 3D maps and keeping them up-to-date using existing technologies, such as laser scanners, is expensive. This paper proposes and implements a novel approach to generate 2.5D (otherwise known as 3D level-of-detail (LOD) 1) maps for free using Global Navigation Satellite Systems (GNSS) signals, which are globally available and are blocked only by obstacles between the satellites and the receivers. This enables us to find the patterns of GNSS signal availability and create 3D maps. The paper applies algorithms to GNSS signal strength patterns based on a boot-strapped technique that iteratively trains the signal classifiers while generating the map. Results of the proposed technique demonstrate the ability to create 3D maps using automatically processed GNSS data. The results show that the third dimension, i.e. height of the buildings, can be estimated with below 5 metre accuracy, which is the benchmark recommended by the CityGML  standard.
\end{abstract}

\maketitle

\section{Introduction}
3D maps are increasingly used in a number of applications, such as urban analytics and decision support systems \parencite{biljecki2015applications,lin2009virtual,dollner2006virtual}, several location-based services including navigation \parencite{verbree2007positioning,groves2012intelligent}, autonomous vehicles \parencite{levinson2011towards}, path planning, map-matching for more reliable obstacle avoidance within drone navigation \parencite{floreano2015science},the safety requirements of some emergency services \parencite{kolbe2008citygml,pu2005evacuation}. However, there are still challenges in creating and keeping them up-to-date. Existing specialist technologies for creating these maps, including airborne photogrammetry \parencite{suveg2004reconstruction} and airborne and ground-based laser scanning \parencite{baltsavias1999airborne, pu2006automatic} are relatively expensive. These costs are not one-off; they must be repeated regularly to keep maps up-to-date. 

Alternative low-cost approaches automate the process of using localised knowledge through data-mining of administrative data \parencite{yin2008generating,biljecki2017generating} or Volunteered Geographic Information (VGI) \parencite{goetz2013towards}. However, data-mining is limited by the coverage and temporal accuracy of the data source, and the quality of VGI generated maps cannot be taken for granted \parencite{fan2014quality,salk2016assessing}. Computer vision techniques (e.g. Structure from Motion algorithms) have been used to create 3D maps from building photographs for an immaterial cost  \parencite{snavely2008modeling}, but these depend on the availability of photographs, which restricts coverage to only the most significant locations or requires deliberate data collection.

This paper proposes the use of automated collection of Global Navigation Satellite System (GNSS) data to find GNSS signal patterns and to create 3D maps with level-of-detail 1 (also known as 2.5D). GNSS signal patterns can be used to create 3D maps because the signals can be blocked or affected only when they interact with the environment and obstacles such as buildings. The advantage of GNSS data is that smartphones already collect the required data to facilitate their location-based services. This allows volunteers to passively contribute VGI by sharing their data. The use of passive contributions of GNSS data has been shown to improve the quality of 2D VGI maps beyond active contributions \parencite{basiri2016using}.   GNSS technology is well-suited to a VGI project as it is a familiar technology often used in such contexts \parencite{huang2013evaluation}; globally available and free-to-use \parencite{langley2017introduction}; and easily accessible given the high penetration of GNSS-enabled smartphones \parencite{ESAuser}. Furthermore, collecting data is straightforward as smartphones running the Android operating system (currently comprising an 85\% market share of new phones sold globally \parencite{IDC}) have allowed access to GNSS raw data since version 7.0 introduced in 2016 \parencite{banville2016precision}. 

Section \ref{gnss} describes how a system would work in practice. A GNSS signal is transmitted from a satellite at a known location (based on the signal’s timestamp) to a receiver with an estimated position (either via GNSS or another navigation solution). If the signal is Line-Of-Sight (LOS) or multipath, it may provide evidence that there are no intervening objects, and conversely non-line-of-sight (NLOS) or blocked signals indicate an interaction with one or more obstacles. These data are processed by a map building algorithm to generate the map, as discussed in section \ref{algo}. 

Classifying GNSS signals between LOS (or multipath) and NLOS is a difficult problem even with specialised equipment and software, but especially for smartphone receivers \parencite{groves2013portfolio}. In this regard this paper proposes a GNSS-based map generating algorithm using the theoretical context of computer vision techniques (section \ref{vision}), which help to measure and assure the achievable quality, e.g. accuracy. It is important to investigate if a bias in the signal classifier can affect map accuracy; and also how the uncertainty due to the signal classifier can be reproduced in the produced map. GNSS mapping works published to-date, as described in section \ref{vision}, demonstrate different proposed algorithms but have not yet addressed these points. Their results \parencite{swinford2005building,kim2008localization,weissman20132,irish2014probabilistic,irish2014belief,isaacs2014bayesian,rodrigues2019extracting} rely on ad hoc justifications for the chosen parameters or use non-generalisable training data-sets of labelled signals , which is addressed in this paper..

We introduce a new mapping algorithm (section \ref{algo}) which is designed to be robust to bias in the signal classifier, and to provide principled estimates of the uncertainty in the map output.  
We apply the algorithm to empirical data collected in a range of urban environments within London and show that it performs consistently across a wide range of initial signal classifiers (section \ref{implementation}).

This paper is structured as follows: Section \ref{gnss} describes the properties of GNSS required to understand the remainder of the paper. Section \ref{vision} discusses 3D mapping and computer vision-based algorithms. Section \ref{algo} proposes the 3D mapping algorithm using GNSS and section \ref{implementation} implements the algorithm and tests its performance. Section \ref{discussion} discusses how the algorithm could be scaled up to a system built upon contribution of data from volunteers at a city-wide scale.

\section{Overview of GNSS} \label{gnss}
At any time, GNSS receivers may receive multiple radio signals transmitted from orbiting navigation satellites. Guides to the principles of a GNSS system are widely available, for example the book "Gnss data processing. volume 1: Fundamentals and algorithms" by \textcite{subirana2013gnss}. It describes the GNSS signal as comprising a ranging code, also known as the pseudo-random-noise (PRN) code, and navigation data. The receiver uses the ranging code to identify the transmitting satellite and calculate the signal travel time from satellite to receiver, which is multiplied by the speed of light to provide the apparent distance, known as the pseudorange \parencite*[65]{subirana2013gnss}. The navigation data provides information on the satellite location along with supplementary information \parencite*[18-37]{subirana2013gnss}. Knowledge of satellite locations and pseudoranges allows the receiver to calculate its position and time if enough satellites (usually 4 or more) are received \parencite*[18-37]{subirana2013gnss}. 

The GNSS signals are used in 3 parts of the proposed 3D mapping system. Firstly, the GNSS position solution can be used to estimate the receiver location and timestamp the signals, although this is optional if other location estimates, including map or street view matching from the user inputs, are available. Likewise the receiver's own clock can be used if the time can not be calculated. Secondly, the timestamp is used to obtain satellite positions from an authoritative reference such as the Navigation Support Office at the European Space Agency \parencite{mayer2019esa}. Using an external reference, as opposed to transmitted navigation data, provides the location of all satellites including those without received signals. Lastly, the signal can be classified in relation to its propagation path, as detailed below, using signal features such as the pseudorange and the Carrier-to-Noise ratio (\CN). The \CN\ is not directly part of the transmitted information but is a measurement by the receiver of the strength of the received signal.

\begin{figure}[h]
\centering
\begin{subfigure}{0.4\textwidth}
\includegraphics[width=\linewidth]{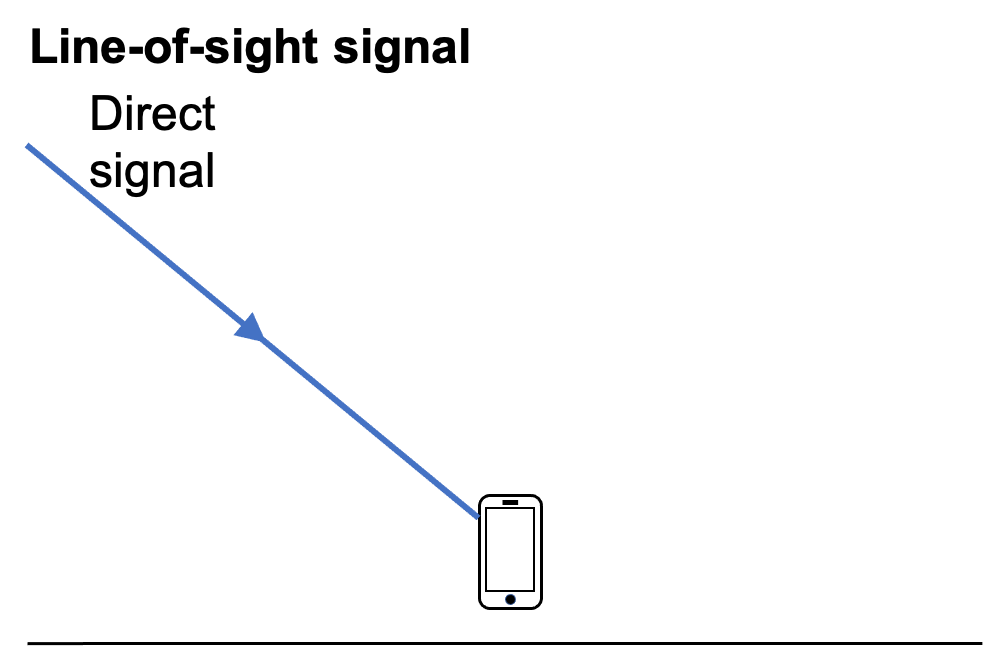}
\end{subfigure}
\begin{subfigure}{0.4\textwidth}
\includegraphics[width=\linewidth]{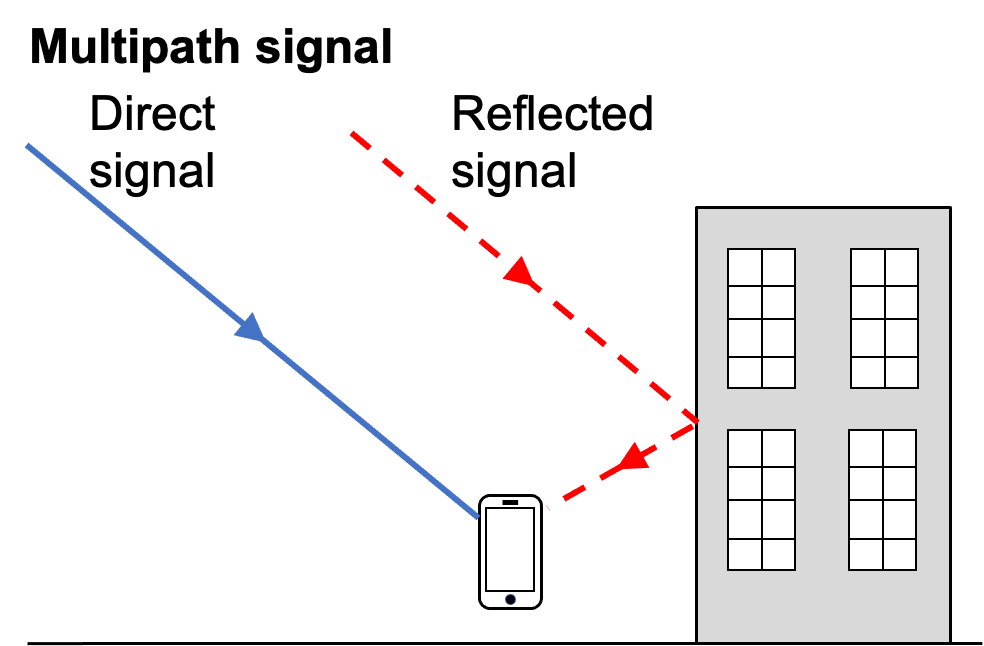}
\end{subfigure}
\vfill
\begin{subfigure}{0.4\textwidth}
\includegraphics[width=\linewidth]{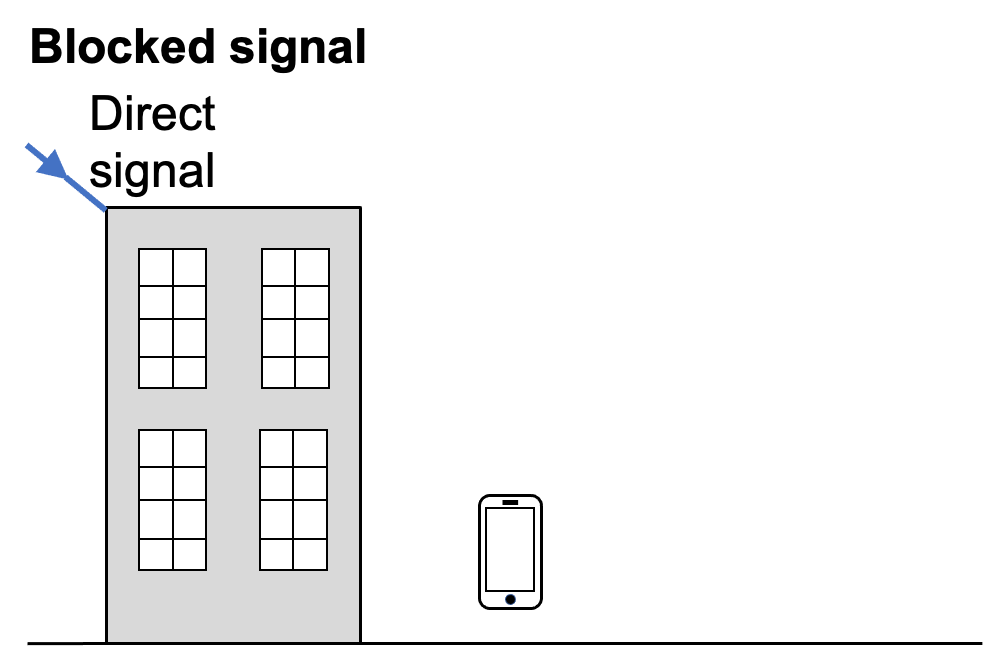}
\end{subfigure}
\begin{subfigure}{0.4\textwidth}
\includegraphics[width=\linewidth]{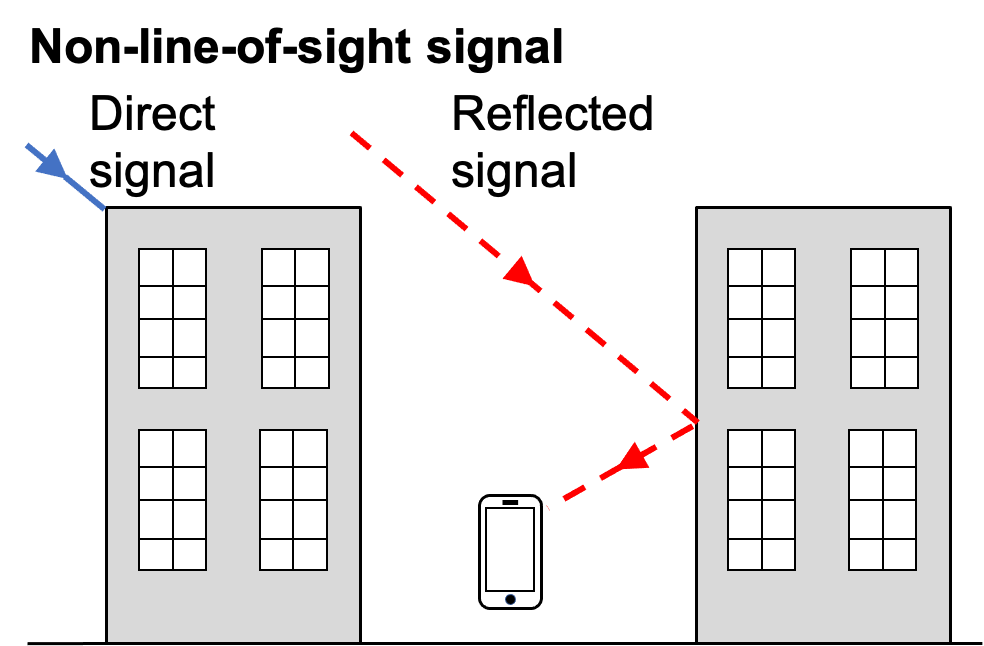}
\end{subfigure}
\caption{GNSS signal types}
\label{fig:signaltypes}
\end{figure}

Like other radio-waves, GNSS signals can propagate to the receiver via a number of different paths. Each path may involve different interactions with the environment, and each signal can be made up of multiple path components. As illustrated in figure \ref{fig:signaltypes}, this leads to classifying GNSS signals into 4 types:
\begin{itemize}
    \item if the signal comprises one path without intervening obstacles, it is line-of-sight (LOS) (see figure \ref{fig:signaltypes},~top-left), 
    \item if the signal comprises one path that has been reflected or diffracted (the bending of a signal around an edge) by obstacles, it is known as non-line-of-sight (NLOS) (see figure \ref{fig:signaltypes},~bottom-right),
    \item if the signal comprises multiple components, which may include a LOS component or be entirely NLOS, it is “multipath” (see figure \ref{fig:signaltypes},~top-right),
    \item if no signal is received at all it is blocked (see figure \ref{fig:signaltypes},~bottom-left).
\end{itemize}
\FloatBarrier
For the purpose of 3D mapping, we must distinguish between a LOS component being present (LOS or multipath with a LOS component) or absent (blocked, NLOS or multipath without a LOS component). The identification of NLOS and multipath signals is the basis of many approaches to improving GNSS positioning accuracy in urban locations, however reliably distinguishing between LOS and NLOS signal components is a challenging problem  \parencite{bressler2016gnss}. Smartphone receivers are particularly prone to receiving NLOS signals because their linearly-polarised antenna are equally sensitive to LOS and reflected signals, whereas professional or traditional geodetic receivers are less sensitive to reflected signals \parencite{chen2012antennas}. At the same time, there are few techniques to identify NLOS signals using a smartphone, because of hardware limitations relating to cost, size and power consumption \parencite{groves2013portfolio}. Three main types of approach have been identified as suitable for smartphones, as detailed below.

The first set of approaches classify signals based on the environment, either using satellite elevation on the principle that a low elevation signals are more likely to be blocked by obstacles \parencite{yozevitch2016robust}, or using a 3D city model as part of the position solution to jointly estimate position and identify NLOS signals \parencite{bourdeau2012tight,obst2012urban,peyraud2013non,yozevitch2014gnss,wang2013gnss}. This is the inverse problem to creating a 3D map, hence such methods are not directly applicable to 3D map creation. 

The second set of approaches identify signals that are likely to be NLOS because the indirect path travelled makes their observed measurements, primarily pseudorange, inconsistent with other measurements. These include calculating position solutions using different combinations of GNSS signals at the same epoch \parencite{groves2013height,jiang2011multi}; comparing measurements from other sensors in an integrated positioning system \parencite{soloviev2009use}; and comparing measurements against prior position estimates using Kalman filters \parencite{groves2013portfolio}. \textcite{jiang2011multi} showed that consistency checking has limited sensitivity for a typical (single-frequency) smartphone receiver as errors due to the atmosphere's effect can be as large as typical pseudorange errors. Furthermore, they depend on the accuracy of other measurements, and in complex urban environments with many reflected signals consistent pseudoranges can be generated by different subsets of NLOS observations \parencite{jiang2012gnss}. 

The third set of approaches use the carrier-to-noise ratio or derived measures to classify the signal. These approaches are based upon the signal attenuating when interacting with obstacles, however the effects of signal propagation are well-known to be more complex, as described by \textcite{molisch2012wireless}. Reflections that cause NLOS signals vary depend on the reflecting surface and the angle of reflection. The signal can be scattered on reflection from a rough surface, leading to a weaker signal in several directions \parencite*[64-66]{molisch2012wireless}, or specular reflection can be generated by a smooth surface, leading to a strong signal in a particular direction \parencite*[49-51]{molisch2012wireless}. Likewise, the effect on signal-strength depends on the degree of diffraction \parencite*[54-63]{molisch2012wireless}. Furthermore multipath signals have a larger variance in their \CN\ because of small-scale fading, the effect of relative phase on superposition of combining signals \parencite*[27-29]{molisch2012wireless}.
As a result, it is possible for a specular reflection to be recorded as strong as or even stronger than a LOS signal \parencite{groves2013portfolio}, and a GNSS signal can still be received at up to 5 degrees of diffraction \parencite{bradbury2007prediction}. Because of this, a smartphone GNSS receiver in an urban environment may produce highly overlapping \CN\ distributions for LOS and NLOS signals \parencite{wang2015smartphone} and \CN\ can be an inaccurate NLOS classifier \parencite{yozevitch2016robust}. 

As a further source of difficulty, the \CN\ can also be affected by non-propagation errors: receiver gain varies by smartphone make, model, and orientation \parencite{chen2012antennas}; satellite transmission power varies across constellations and the individual satellites due to different generations of specification and a decrease in power as satellites age \parencite{steigenberger2018gnss}; GNSS signals can be affected by atmospheric conditions \parencite{kintner2009gnss} as well as features that are temporary or too small to be included on a map, such as road vehicles or foliage on trees; and the human body attenuates signals, as well as any bag or container in which the smartphone may be placed \parencite{bancroft2011gnss}. Many of these different errors cannot be detected or mitigated in a crowd-sourcing situation due to the range of devices and varying methods of collection \parencite{rodrigues2019extracting}. Therefore any carrier-to-noise classifier calibrated under training conditions is likely to underperform when tested in real world scenarios. 

Applying machine learning methods to creating a NLOS classifier by combining features such as \CN, pseudorange, elevation, and Doppler shift has been investigated using decision trees \parencite{yozevitch2016robust}, support vector machines \parencite{hsu2017gnss, xu2020machine}, nearest neighbours and neural networks \parencite{Xu2018gnss}. However evidence for an improvement in accuracy against a untrained Naive Bayesian classifier on \CN\ is mixed:  \textcite{hsu2017gnss} showed an increase in accuracy from 67\% to 75\% and \textcite{yozevitch2016robust} showed a decrease in false positives from 45\% to around 20\%, however other works \parencite{xu2020machine,Xu2018gnss} considering multiple situations showed the simpler classifier had similarly or better accuracy in some situations. It has not been shown that any improvements generalise beyond the original experimental settings.

The highly overlapping \CN\ distributions are suited to a probabilistic classifier. \textcite{irish2014probabilistic} used a Bayes classifier based on a Rician distribution for multipath \CN\ and log-normal distribution for NLOS, which are the canonical forms \parencite{molisch2012wireless}. A quadratic spline form for the probability density function was proposed by \textcite{wang2015smartphone}. This paper proposes a 4-parameter logistic curve \parencite{healy1972statistical} for the classifier form. It is a natural choice for a probabilistic classifier as an extension of logistic regression, as discussed in more detail in section \ref{implementation}. In common with all other approaches mentioned, its performance relies on the fitted parameters and is subject to the same criticisms with respect to performance in a more general setting.

\section{3D mapping as computer vision} \label{vision}

\begin{figure}[ht]
\centering
\includegraphics[width=0.5\linewidth]{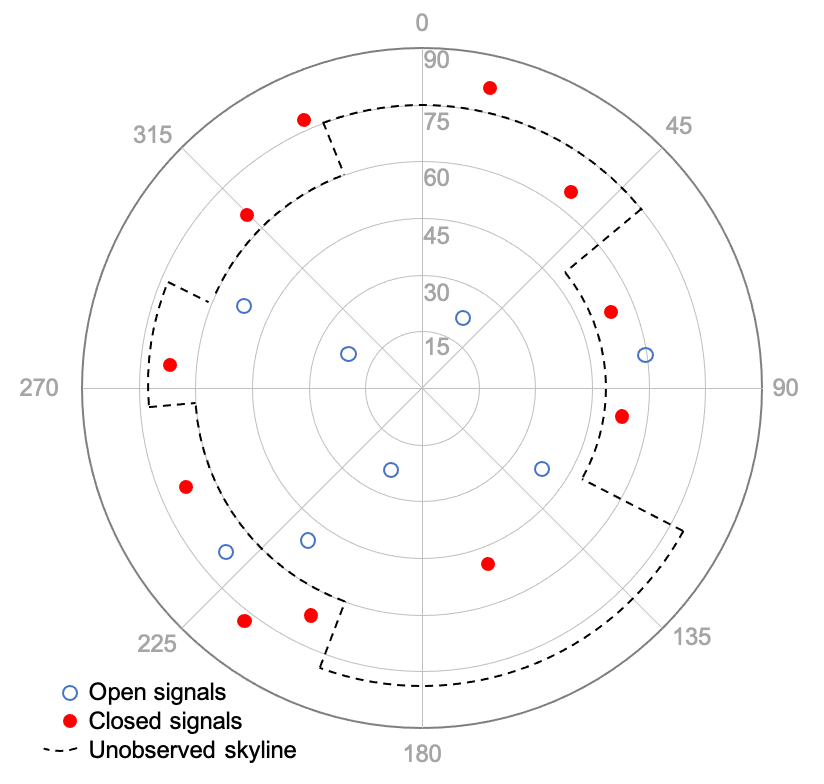}
\caption{Typical skyplot (azimuth-elevation graph) with classified signals}
\label{fig:skyplot}
\end{figure}

Methods for generating 3D models of buildings using existing technologies are surveyed by \textcite{wang20133d} and categorised as either image-based (for example photogrammetry) or range-based (for example LIDAR), or perhaps a fusion of the two. While GNSS is a range-based technology for the purposes of navigation, the measured range is of the satellite and not of the environment we wish to map. However by classifying each signal and considering the satellites' relative positions from the receiver, it is possible to derive an image, known as a skyplot, from the original measurements. This shows the set of satellite observations from a receiver at a single epoch, as illustrated in figure \ref{fig:skyplot}. Similar to a silhouette image, it only shows whether the direct ray to each satellite was open (when the signal is LOS or multipath with a LOS component) or closed (when the signal is blocked, NLOS or multipath without a LOS component). These images can then be used to generate a 3D map through techniques that have similarities to existing image-based techniques. 

\FloatBarrier

The earliest image-based approaches relied on knowing the position and calibrating parameters that produced each image, known as the "camera poses" \parencite{cyganek2011introduction},  and primarily used multi-view stereo (MVS) correspondence \parencite{furukawa2015multi}, where stereographic correspondence determines depth across a collection of overlapping images from multiple viewpoints. More recent approaches use feature identification and matching across images, for example to recover camera pose \parencite{remondino2006image}, however, a camera image is a complete set of observations across a lens’ view, whereas a skyplot is an extremely sparse set of observations, which makes it impossible to identify common features between two images. As common features cannot be identified between GNSS images, a set of MVS techniques apply which use a negative deduction process: assuming the 3D-map has a particular shape, checking how it appears in all the images, and iteratively improving the parts of the object which are inconsistent with the images. Such MVS techniques have been shown to produce sub-millimetre accuracy in lab testing \parencite{seitz2006comparison}. 

GNSS data does not allow a straight-forward application of existing MVS techniques because it is not possible to measure the consistency of the 3D-map to the skyplots with the same precision. This arises because unlike a picture, where parts of images can be compared by several continuous features such as hue, intensity, and brightness, the skyplot only provides the outcomes of a binary classifier with limited accuracy, which limits how consistency can be measured. The limitations of GNSS classification have been discussed in detail in section \ref{gnss}. As a second problem, inaccuracy of receiver position may cause the skyplots to be compared to the incorrect part of the 3D-map. The accuracy of the receiver position may vary depending on how the position is estimated: GNSS position fixes are a natural automatic positioning mechanism, but horizontal errors in built-up urban locations may be up to 50 metres for a smartphone \parencite{wang2015smartphone}; the receiver estimate could also be provided by the user, leading to a semi-automated process subject to existing questions of volunteered data quality \parencite{fan2014quality}. 

While these problems may limit the detail of the produced 3D-map, GNSS mapping is an alternative to existing 3D-map creation at the scale of a city, which historically utilises airborne methods focused on coarse modelling with geometrically simple building and roof structures \parencite{zebedin2006towards,lafarge2012creating}. CityGML, the Open Geospatial Consortium standard for 3D models, defines 5 increasing Levels Of Detail (LOD) with  LOD 1 representing each building as an extruded block with flat roof \parencite{kolbe2009representing}. LOD 1 is sufficient for many 3D mapping applications \parencite{biljecki2015applications,biljecki2016variants} and is often a target for large scale 3D-map production \parencite{dukai2019multi,girindran2020reliable}.

    \begin{table}[ht]
    \caption{GNSS mapping algorithms}
    \setlength\extrarowheight{6pt}
    \hyphenpenalty=10000
    \footnotesize
    \rotatebox{90}{
    \begin{tabularx}{1.45\textwidth}{
    >{\hsize=0.75\hsize}X
    >{\hsize=0.75\hsize}X
    >{\hsize=1.65\hsize}X
    >{\hsize=0.75\hsize}X
    >{\hsize=0.75\hsize}X
    >{\hsize=1.85\hsize}X
    >{\hsize=0.75\hsize}X
    >{\hsize=0.75\hsize}X
    }
    \hline
    Technique & Scene representation & Photo-consistency & Visibility model & Shape priors & Reconstruction algorithm & Initialisation requirement & LOD\\
    \hline\hline
    \textcite{swinford2005building} 
    & Voxel – 1 metre cubes
    & N/A. NLOS measurements were ignored 
    & Not used 
    & 2.5D mapping (columns of occupied voxels must have no gaps) 
    & Open signals used to progressively remove voxels, as a form of space-carving \parencite{kutulakos1999theory} 
    & 2D map 
    & 1
    \\[1cm]
    \textcite{kim2008localization} 
    & Voxel – unknown size
    & N/A. LOS measurements ignored
    & Not used 
    & 2.5D mapping
    & Counts of NLOS measurements used to determine 2D footprint, followed by cost-thresholding of counts to determine height 
    & Bounding box 
    & 1
    \\[1cm]
    \textcite{weissman20132}
    & Voxel – unknown  size                                                          & Hinge-loss function for building height, using height of incorrectly classified signals intersecting a vertical column of voxels
    & Not used
    & 2.5D mapping
    & Counts of NLOS measurements used to determine 2D footprint, building height estimated by minimising hinge-loss. 
    & Bounding box
    & 1
    \\[1cm]
    \textcite{irish2014probabilistic,irish2014belief}
    & Voxel – 4 metre cubes
    & Naive Bayesian probability of a voxel being occupied given set of intersecting signals and a \CN\ probabilistic classifier
    & Probability of intervening voxels being unoccupied & None.
    & Calculates posterior probabilities for each voxel being occupied (modelling as a factor graph and using a sum-product algorithm). Similar technique to probabilistic space carving \parencite{broadhurst2001probabilistic}. Extended to allow for gaussian noise in receiver location \parencite{irish2014belief}. 
    & Bounding box
    & N/A - surface extraction unspecified
    \\[1cm]
    \textcite{isaacs2014bayesian}
    & Voxel – unknown size                                 
    & Naive Bayesian probability of a voxel being occupied given set of intersecting signals and a \CN\ probabilistic classifier 
    & Not used
    & None.                                                         
    & Calculates probabilities for each voxel through an online learning algorithm that applies the update to all voxels intersected by the signal, by stepping from their current value toward the photo-consistency measure.                      
    & Bounding box. Initialised using Irish et al. method
    & N/A - surface extraction unspecified
    \\[1cm]
    \textcite{rodrigues2019extracting}
    & Voxel – 4 metre cubes                                 
    & Binary classifer based on \CN\ features of the set of intersecting signals, with several machine learning methods tried.
    & Not used
    & None.                                                         
    & Applies a trained classifier to each voxel, with intersecting signals weighted to allow uncertainty in receiver location  & Bounding box
    & 1
    \\[1cm]
    4PL-B (proposed in section \ref{algo})
    & 2D surface mesh
    & The likelihood of a building’s classifier given the observed set of signals (classified by signal features)	
    & Not used
    & 2.5D mapping. All buildings have a single height parameter
    & On a building-by-building basis: maximum likelihood of the associated four-parameter logistic regression with intersection height as the independent variable 
    & 2D map
    & 1\\[1cm]
    \hline
    \end{tabularx}
    }
    \label{table:algos}
    \end{table}
\FloatBarrier

To understand these limitations in practice, table \ref{table:algos} categorises existing GNSS-mapping techniques and the algorithm proposed in this paper (section \ref{algo}) against a taxonomy of MVS techniques introduced by \textcite{seitz2006comparison}. This categorises an MVS algorithm by six fundamental properties: scene representation, photo-consistency measure, visibility model, shape prior, reconstruction algorithm, and initialisation requirements. As an additional category, we add the LOD of the produced map.
\begin{enumerate}
    \item Scene representation is how the object geometry is represented. Primary types are voxel (a 3D grid of pixels) or a 2D surface mesh. Many reconstruction algorithms require a particular representation.
    \item Photo-consistency is a measure of the consistency between images to determine whether an object reconstruction (which defines the projected images) is consistent with a set of images. 
    \item	Visibility model is the method by which images are selected when evaluating photo-consistency measures. It is important to exclude images where the object is occluded.
    \item Shape priors guide the reconstruction algorithm to produce desired object characteristics such as smoothness and maximal or minimal surface area or volume. 
	\item Reconstruction algorithm is the method of producing the reconstruction. Non-feature based methods typically work by minimising inconsistency cost either i) point-by-point (cost function defined for each scene position and surface extracted based on a cost threshold); or ii) globally (cost function defined for an object surface) 
	\item The initialisation requirement provides constraints on scene geometry. Many algorithms require constraints (for example a bounding box) to function, and may obtain different solutions depending on their initialisation. 
\end{enumerate}

Existing GNSS mapping algorithms seem to be relatively simplistic compared to existing MVS algorithms. They are primarily distinguished by their varying approaches to photo-consistency, which reflects the limitations discussed above. However, none of the approaches to date are satisfactory for generating even an LOD 1 map. This is because the measurements of photo-consistency either ignore inconsistent classifications \parencite{swinford2005building,kim2008localization}, or rely on a pre-specified signal strength classifier to turn signal strength observations into a cost \parencite{irish2014probabilistic,irish2014belief,isaacs2014bayesian,weissman20132,rodrigues2019extracting}, taking for granted that the classifier model is unbiased. The probabilistic approaches of \textcite{irish2014probabilistic} and \textcite{isaacs2014bayesian} are potentially more robust because they effectively give a higher weighting to observations with a more certain classification, however the effect of model misspecification has not been investigated. 

The algorithms also do not specify how uncertainty in the photo-consistency measure (due to signal classification error and location error) should be reflected in the reconstruction algorithm.  We emphasise that the probabilistic approaches do not address this, as all posterior probabilities will converge to 0 or 1 with a large enough data set.  The probabilistic approaches do not specify how a surface should be extracted from the voxel probabilities, but probabilistic space carving is typically used with a cost-minimisation approach, where cost would include negative likelihood as well as potential shape priors. It is possible to reflect uncertainty through considering the sensitivity of cost to changes in the map, however cost sensitivity is also a function of the chosen classifier and size of the data set , making such approaches problematic. The algorithm we introduce in the next section seeks to address both problems of bias and uncertainty.

\section{GNSS-mapping algorithm} \label{algo}
\begin{figure}[h]
\centering
\includegraphics[width=0.5\linewidth]{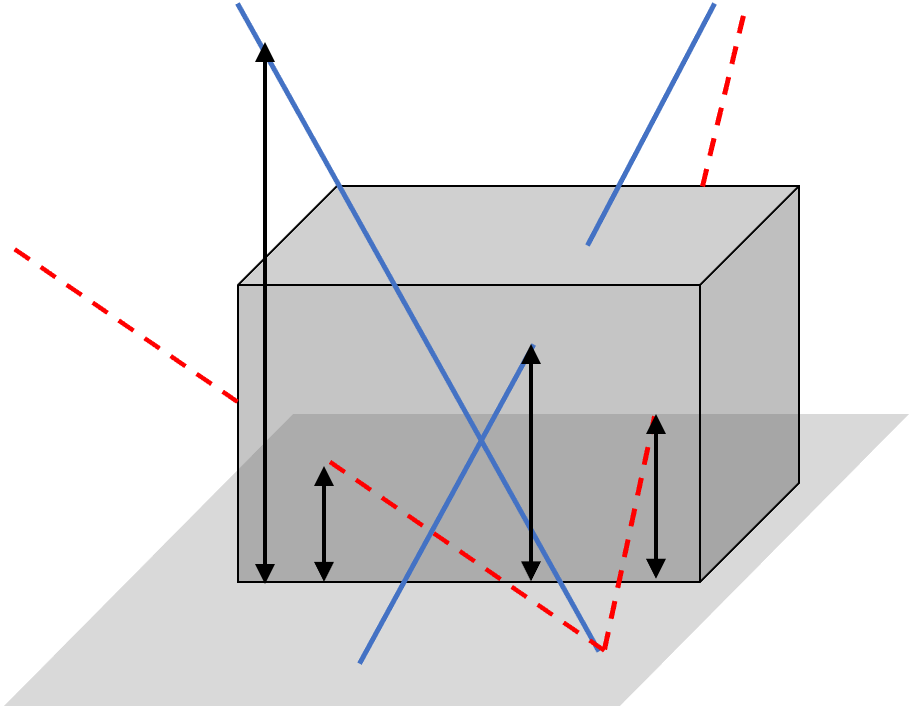}
\caption{Example of building data generated for 2.5D mapping from footprint intersection heights and signal classifications}
\label{fig:algosystem}
\end{figure}
Our proposed GNSS mapping technique directly generates a 3D map by using it as a signal classifier and solving the inversion problem given the observed signal data, similar to the MVS approach known as statistical inverse ray tracing \parencite{liu2011complete}. The technique is applied to a single building at a time, using an existing 2D map to provide a building footprint. The set of GNSS observations that intersect the 2D footprint form a dataset of (intersection height, signal features) pairs, as illustrated in figure \ref{fig:algosystem}. The 3D map is extruded from the 2D footprint with the building height as an unknown parameter. The technique uses a classifier based on signal features to label the GNSS signals and the labelled signals are used as data to estimate the building height, as detailed below. This process is then reversed with signals labelled as open or closed depending on whether they intersect the 3D building, and these relabelled signals are used to train the signal classifier.  This process iterates to improve the building height estimates. Table \ref{table:algos} categorises the algorithm alongside existing approaches. 

The proposed algorithm has three novel aspects relating to the photo-consistency problem:

\begin{itemize}
    \item Firstly, it considers intersection height as a probabilistic classifier, using a four-parameter logistic regression (4PL) as defined below as the classifier form for each building. This leads to a natural interpretation of height uncertainty from the certainty of the classifier. 
    \item Secondly, rather than generating the map by maximising its accuracy as a classifier, which creates unwanted dependence on the bias of the signal feature classifier, it relates the map classifier parameters to the physical map, based on our understanding of the typical measurement errors of the observations. 
    \item Thirdly, once an initial map is generated, it is used to reclassify the signals and refit the signal feature classifier, before repeating in a bootstrapping approach. This takes advantage of shape priors to improve the map beyond observed data. 
\end{itemize}
These aspects are discussed further following the details of the technique, which is set out as the following pseudo-code \ref{algo:4plb}. For simplicity it assumes the only signal feature used for classification is the carrier-to-noise ratio, however any signal classifier could be used, as long as it is class-conditionally independent of height, and  section \ref{gnss} contains more detail on potential classifiers. 

\begin{algorithm}[H]
\SetKwFunction{SignalClassifier}{SignalClassifier}\SetKwFunction{FourPL}{4PL}
\SetKwProg{Fn}{function}{:}{}
\SetKwInOut{Input}{input}\SetKwInOut{Output}{output}

\Fn{\SignalClassifier(param,ss)}{\tcc{signal strength classifier, with heuristic initial parameters}
\leIf{ss=n/a}{return 0}{return $\Pr(ss)$}
}  
\BlankLine
\Fn{\FourPL((a,b,c,d),h)}{\tcc{four parameter logistic regression associated with the building, with intersection height as the independent feature used to classify signals}return $\Pr(h)$\;}
\BlankLine
\Input{A set $X$ of observation tuples $(y,ss,h)$ for a building
\linebreak \indent\indent $y$: signal label (initially blank)
\linebreak \indent\indent $ss$: \CN\ (n/a if blocked)
\linebreak \indent\indent $h$: intersection height}
\Output{Parameters $(a,b,c,d)$ of \FourPL
\linebreak \indent\indent $c +1.5/b$: height point estimate
\linebreak \indent\indent $(c,c+3/b)$: height range estimate}
\BlankLine
\BlankLine

\Repeat{$(a,b,c,d)$ have converged}{
    \lFor{$(y,ss,h)\in X$}{
         $y \longleftarrow \begin{cases} 1 & SignalClassifier(param,ss) > 0.5  \\
         0 &  o'wise
        \end{cases}$
        }
    
    $(a,b,c,d) \longleftarrow \text{maximum likelihood estimates for } \FourPL \text{ given } X$
    
    \lFor{$(y,ss,h)\in X$}{
        $y \longleftarrow \begin{cases} 1 & h>c  \\
         0 &  o'wise
        \end{cases}$
        }
        
    $param \longleftarrow \text{maximum likelihood estimates for } \SignalClassifier \text{ given }X$
    }
\caption{4PL-B algorithm}
\label{algo:4plb}
\end{algorithm}

A 4PL extends logistic regression to consider asymptotic probabilities other than 0 or 1\parencite{healy1972statistical}. It has the following univariate form: 
\[\Pr(x)=d+ \frac{a-d}{1+\exp(-b(x-c))}\]
Where $a$ and $d$ are the upper and lower asymptotes, respectively, and $b$ and $c$ are the standard linear coefficients in $x$. The 3D mapping algorithm uses the maximum likelihood estimates of the 4PL parameters, which can usually be obtained through standard optimisation methods although the EM algorithm has been proposed as a superior alternative  \parencite{dinse2011algorithm}. Gradient descent methods \parencite{ruder2016overview} can also be used to optimise the parameters, which raises the possibility of using stochastic gradient descent on the classifiers and performing batch-updates on each classifier in turn, in order to allow the 3D map to be updated over time as new data is collected. 

\begin{figure}[h]
\centering
\includegraphics[width=0.5\linewidth]{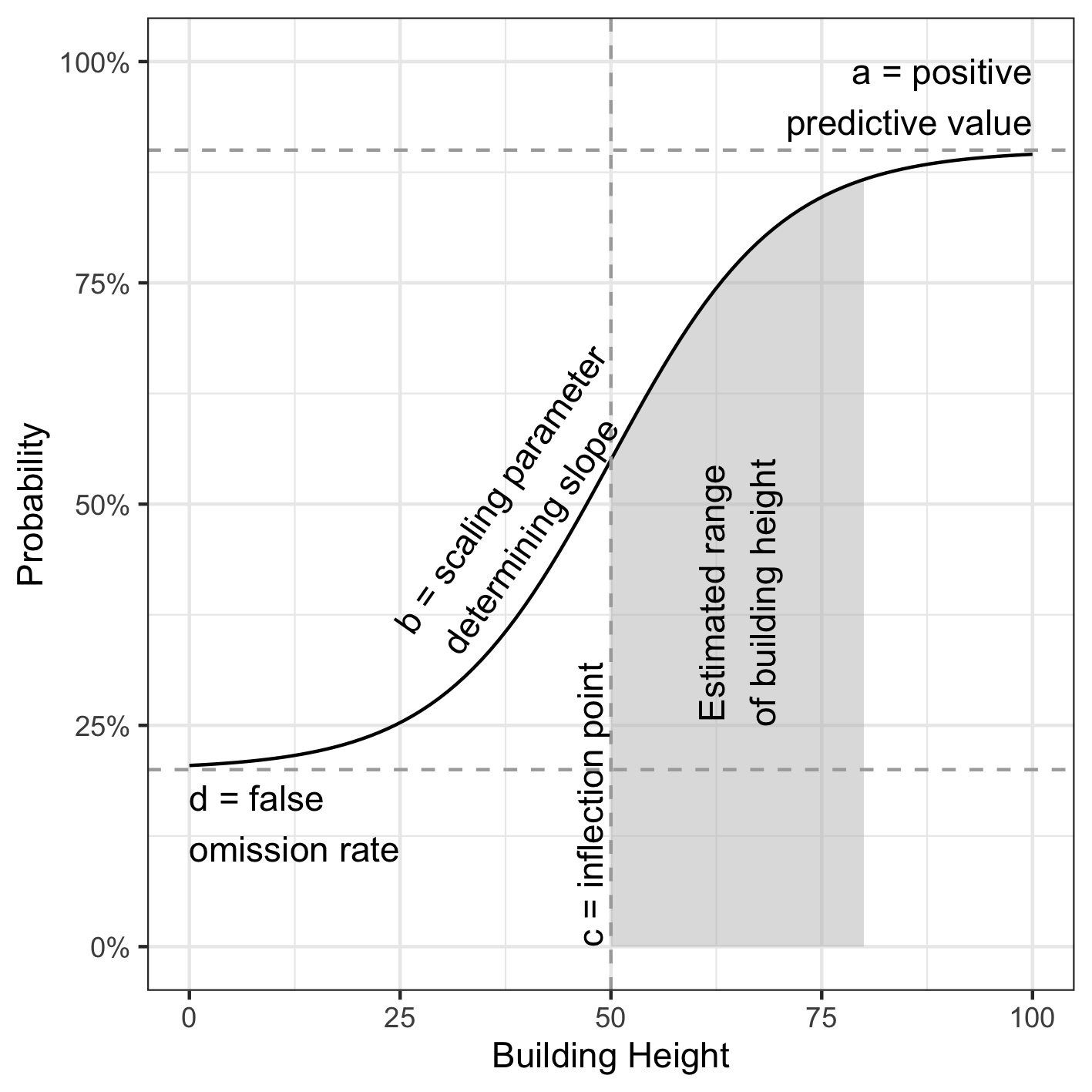}
\caption{Four-parameter logistic regression}
\label{fig:4pl}
\end{figure}

The use of the 4PL is integral to the proposed technique as the parameters have a physical interpretation relating to the observation errors, as illustrated in figure \ref{fig:4pl}. Without these errors, an idealised height classifier would be a step function, with zero probability of open signals for intersections below building height and probability of one above building height. 

The 4PL asymptotes reflect signal classification error, where  $a$ is the positive predictive value, or precision, of the signal classifier and $d$ is the false omission rate (equal to 1 minus the negative predicative value). It is important to clarify that these are primarily the error rates of the signal classifier which has noisily labelled the data, and not the error rates for the map classifier against the true unknown labels. 

The scaling parameter $b$ relates to the uncertainties in building height, for reasons including unmodelled effects such as diffraction effects and the complexities of building roofline, and measurement errors such as incorrect intersection height due to receiver location errors. A gentle slope indicates a lower level of certainty in building height as the effect of height on signal class is prolonged, whereas a steep slope indicates more certainty due to an abrupt height effect. 

The inflection point $c$, relates to the unknown building height. The $b$ and $c$ parameters generate a height estimate of $c+1.5/b$ with a range between $(c, c+3/b)$ as explained below.

\FloatBarrier

The rationale for this approach is that the GNSS received signal strength and intersection height can be treated as class-conditionally independent, with the exception of diffraction effects and height errors. Class-conditionally independent means that within a signal classification (open or closed) the signal strength is independent of the intersection height: for open signals (where the signal intersection height is above the building height) there is a LOS component which does not interact with any obstacles and this is unaffected by intersection height; likewise, for closed signals (where the signal intersection height is below the building height), the LOS component has been blocked and this is unaffected by the intersection height. 

This is a generalisation: the contribution of multipath and NLOS components as well as any effects of unmapped objects on LOS components will depend on position but with a reasonable spatial distribution of observations around the building the dependence on height is presumed to be ignorable; the algorithm also assumes that the blockage of signals by other buildings (which will have a height dependence) does not have an effect, but this is a similar approach to other MVS algorithms, which correct for this problem using a visibility model rather than in the photo-consistency measure; the remaining dependence comes from diffraction effects and height errors, which are explicitly accounted for in the algorithm.

When diffraction effects and height errors are negligible, the class probability (which is measuring the consistency of the signal strength classifier and map classifier) is independent of intersection height, as represented by the asymptotes of the 4PL with zero gradient. As the intersection height approaches the building height from either direction, a growing proportion of signals may be misclassified by the map due to height errors. This occurs when the intersection height is between the modelled building height (which is modelled as uniform across the building) and the actual building height, which varies across the building with roofline complexity, or conversely, when the actual building height is between the measured and actual intersection height due to receiver location errors. 

These effects lead to a slope between the asymptotes instead of a step, however the effect of intersection height errors should be symmetric, and therefore not affect $c$, the inflection point of the 4PL. The effect of non-uniform building height will shift $c$ to the average building height across the observations, which should be a good approximation of the true average building height with spatial diversity of observations. 

On the other hand, diffraction effects depend on the degree of diffraction of the signal which depends on both the distance of the receiver as well as the difference between building height and intersection height. As a simplification, the algorithm assumes that the height difference is a proxy for the degree of diffraction, and as this decreases the signal attenuation lessens, leading to an increasing number of LOS classifications. 

This is an asymmetric effect, as it only has an effect when the intersection height is below the building height. This has the effect of shifting the inflection point $c$ below the true building height, therefore the building height should be above the inflection point and towards the upper asymptote. Given the uncertainty of the relative magnitude of intersection height error and diffraction effect, we conservatively suggest a point estimate of $c+1.5/b$ and range between $(c, c+3/b)$, which represents the 50\% to 95\% percentile of the distance between the asymptotes.

It follows from this formulation that our estimate does not impose any additional assumptions on the signal strength classification accuracy, whereas a solution provided by a cost-minimisation approach depends on the signal classification model being an unbiased classifier. The proposed approach provides a height estimate that is robust to model bias and also provides a measure of uncertainty in the height due to the underlying measurement uncertainty.

In order to improve the classification accuracy, this paper builds upon the class-conditionally independency of GNSS signal strength and height, and uses a co-training approach to label data more confidently through an iterative process. \textcite{blum1998combining} introduced co-training as a form of disagreement-based semi-supervised learning, which uses a pair of binary classifiers to label a dataset by iteratively applying each classifier and using the most confidently labelled instances as training data for the other classifier. In the straightforward bootstrapping approach we take here, we consider all labelled signals as training data, and stop when the map classifier parameters converge. This does not require the two classifiers to have consistent labels, but indicates that the classifiers are class-conditionally independent.

\section{Implementation} \label{implementation}
To verify the accuracy of our proposed algorithm, we estimated the height of a range of buildings in different urban environments in greater London by collecting GNSS observations nearby and using them as inputs into the proposed algorithm. The next subsections describe the experimental setup and algorithm results.

\subsection{Experiment}
\begin{figure}[h]
\centering
\begin{subfigure}{1\textwidth}
\includegraphics[width=1\linewidth]{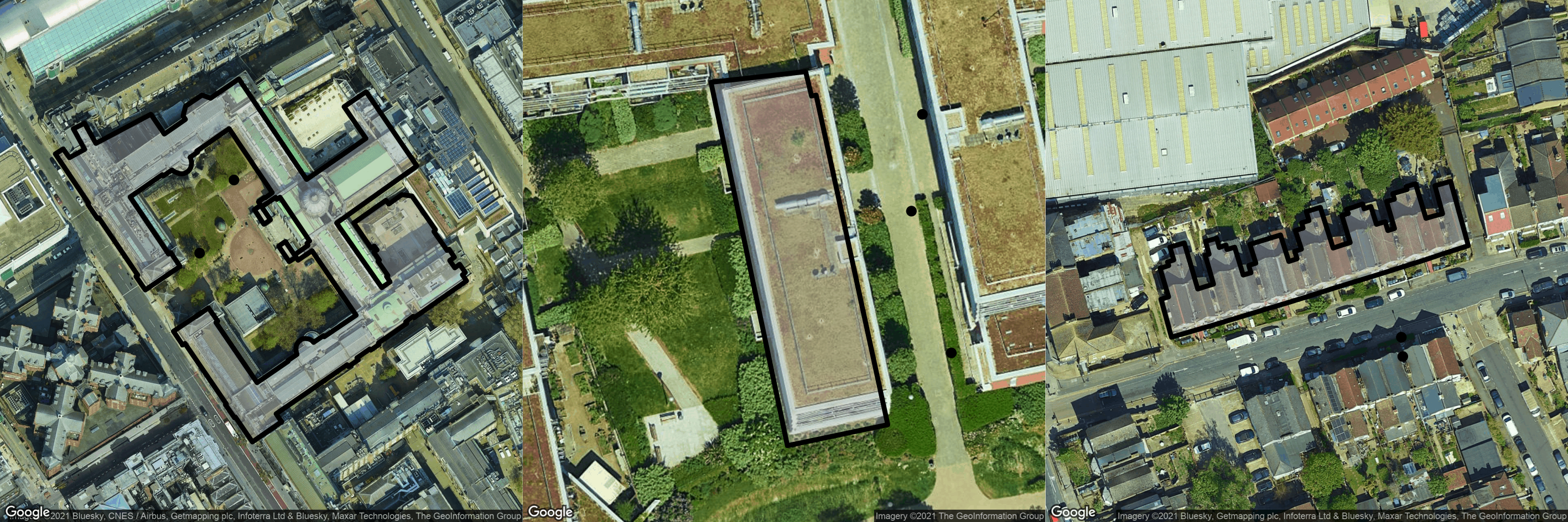}
\end{subfigure}
\vfill
\begin{subfigure}{1\textwidth}
\includegraphics[width=1\linewidth]{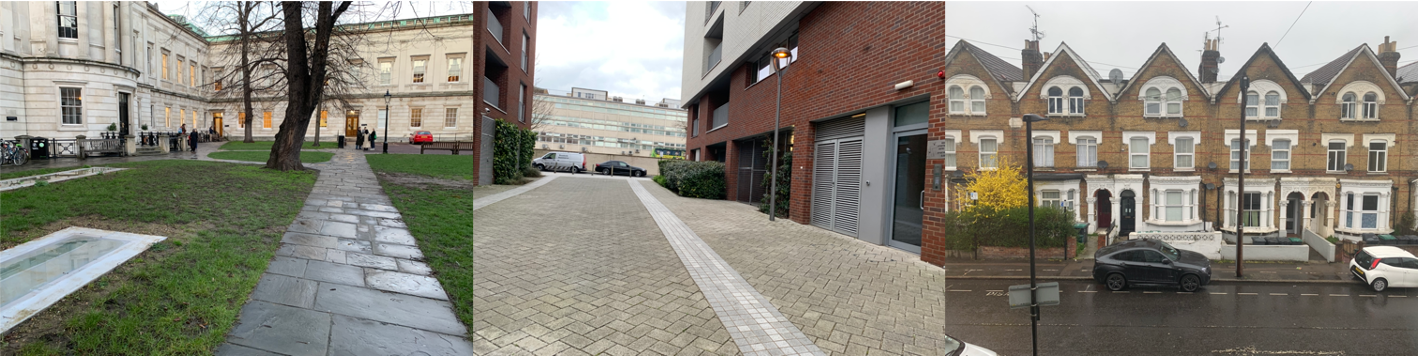}
\end{subfigure}

\caption{Experiment locations above, and photos below. From left to right: UCL, Goodchild Road, Hermitage Road. The outlined polygon is the building footprint according to OS Mastermap, this was used in the algorithm with the background map tile for illustration only. Background map tile: Google, \copyright2021 Bluesky, CNES/Airbus, Getmapping plc, Infoterra Ltd \& Bluesky, Maxar Technologies, The GeoInformation Group}
\label{fig:location}
\end{figure}
 
Urban environments in greater London vary in use, street width, building height and density, and construction material. To compare algorithm performance in different environments, three representative buildings are presented which cover a wide range of building heights, construction materials, and sky visibility(see Figure \ref{fig:location}). The first is the main estate of University College London, the second is a block of flats (Goodchild Road) in an urban canyon, and the third is a terrace of houses (Hermitage Road) in a residential setting. The university estate is a building of significant height and mass, built primarily from stone in the 19th century, with nearby open space allowing spatial diversity of data collection. The block of flats represents modern inner-city construction: it is of moderate height in a high-density location, built within the last 20 years from reinforced concrete with a brick and glass facade, the observations were made from a narrow pedestrianised street with visibility to the sky only at high elevations. The terrace is low density and low height, it is brick-construction from early 20th century, and the observations were made with a restricted collection protocol from the far side of the street to maximise the number of received low elevation signals. 

A Samsung Galaxy 10 running the Android 10.0 operating system was used as the GNSS receiver. It supports the four major GNSS constellations: Global Positioning System (GPS), GLObal NAvigation Satellite System (GLONASS), Beidou, and Galileo. Multi-constellation receivers such as these ones are now standard and embedded in many mobile devices \parencite{ESAuser}, and result in  a device typically having potential LOS to 30 to 50 GNSS satellites above the horizon at all times. As described further in section \ref{discussion}, Android phones allow access to "raw" GNSS data for all signals, along with associated GNSS receiver position solutions. We developed a recording app based on Google’s GNSS Logger \parencite{banville2016precision} that accesses the raw data for the position solution and satellite identifier, used to determine the direct signal path, and the carrier-to-noise density, used to classify the signal. The app also allows manual recording of observer locations.

For each site the receiver was placed in a static location and measurements were recorded as a stream of data measured at 1Hz frequency for a  duration of around 30 minutes. This was undertaken twice at UCL, and 5 and 6 times at the Hermitage Road and Goodchild Road sites respectively, at different locations and times of day to allow the satellite geometries to change. The varying frequency of measurements across sites were to provide similar numbers of intersecting signals across the different sizes and shapes of floorplates. The dataset is summarised in table \ref{table:data}. The observer locations were manually inputted based on a visual determination against a small-scale map, with an estimated location error of below 2 metres. Results are presented in this paper for both the more accurate manually input locations and the automatic GNSS position solutions 

\begin{table}[h]
\centering
\small
\setlength\extrarowheight{6pt}
\caption{Summary of collected data}
\begin{tabularx}{\textwidth}{|
>{\hsize=1\hsize}l
>{\hsize=1.00\hsize}X
>{\hsize=1.00\hsize}X
>{\hsize=1.00\hsize}X
>{\hsize=1.00\hsize}X
>{\hsize=1.00\hsize}X|
}
\hline
\multirow{2}{4em}{\newline Site ID} & \multirow{2}{4em}{Recorded\newline Epochs} & \multicolumn{4}{c|}{Number of signals observed}  \\
\cline{3-6}
& & Recorded & Blocked &  Total &  Intersecting \\
\hline
UCL & 3,558 & 88,633 & 5,284 & 93,917 & 88,106 \\
Goodchild Rd & 8,136 & 227,658 & 21,718 & 249,376 & 95,362\\
Hermitage Rd & 8,350 & 26,094 & 310,618 & 336,712 & 100,884 \\
\hline
\end{tabularx}
\label{table:data}
\end{table}

 For each observation time, all satellite positions were retrieved using information on satellite orbits freely available from the Navigation Support Office at the European Space Agency, which allowed the dataset to be expanded to blocked signals for all receiver-supported constellations. An elevation filter was applied to remove low elevation signals, as these are often blocked by background and unmapped objects. The elevation filter was set at 10 degrees for the UCL and Goodchild Road sites, and 0 degrees for the Hermitage Road site, where the recordings were made from a relative distance and position such that many low elevation signals were received. As a last step, observations with a satellite elevation above 85 degrees were also removed, because at these elevations trigonometry implies errors in the observer’s horizontal position are magnified 11 times or more in calculations of intersection height, making the data unacceptably poor quality. The elevation was calculated with respect to the tangential plane on the WGS84 ellipsoid.

\begin{figure}[h]
\centering
\includegraphics[width=0.9\linewidth]{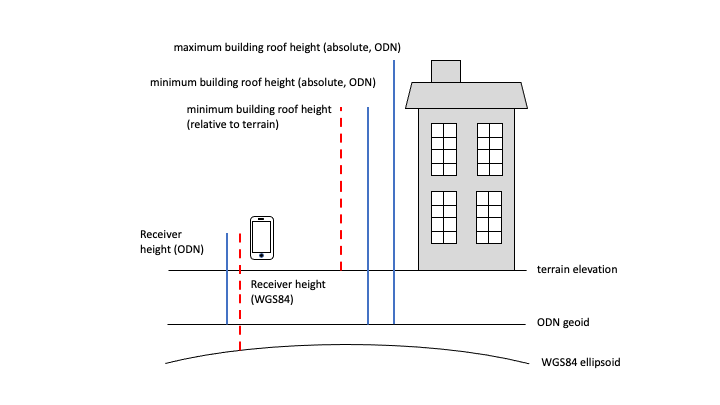}
\caption{Building height definitions}
\label{fig:heightcrs}
\end{figure}

The building footprints were obtained from the Ordnance Survey GB, Great Britain's national mapping agency. Ordnance Survey's MasterMap \parencite{OSMastermap}, provides the benchmark map for the country, in terms of reliability and accuracy. GNSS elevation data is collected by a receiver with reference to the WGS84 ellipsoid. It was converted using accurate transformation grids, as described in \textcite{OSGuide}, to the coordinate reference system used by OS Mastermap: British National Grid for horizontal coordinates and Ordnance Datum Nelwyn (ODN), a local geoid, for heights. Figure \ref{fig:heightcrs} illustrates the conversion between ODN and WGS84 and the relation between absolute building height and relative building height from terrain. The results in this paper are reported as absolute heights with reference to ODN. It also illustrates how building heights in a LOD 1 model depend on the choice of geometric reference: the lowest or highest point of a roof may be used, or any ad-hoc point or average of points in-between \parencite{biljecki2016variants} 

The building heights, shown in table \ref{table:heights}, were obtained from OS MasterMap \parencite{OSHeight} where available, and Google Earth \parencite{googleearth}, and represent the minimum and maximum heights across the buildings, with the UCL building being particularly complex with a large dome and varying levels. For the UCL building, the adopted height in the paper is the height of the Library entrance hall, which we believe is representative of the larger building. While this is a subjective choice, the value of our algorithm lies in its consistent results over a range of starting conditions, and this conclusion holds for any reasonable adopted height for the UCL building. We adopt the midpoint of the height range as the ground truth height.1 Table \ref{table:heights} also shows the indicative ground elevation and relative building heights, determined using the OS Terrain 5 Digital Elevation Model \parencite{OSDTM}.

\begin{table}[h]
\centering
\small
\caption{Building heights}
\setlength\extrarowheight{6pt}
\begin{tabularx}{\textwidth}{|lXXX|}
\hline
Site & UCL & Goodchild Road & Hermitage Road\\
\hline
OS Height (min-max)  & 43.8m - 47.1m  & - & - \\
Google Earth (min-max) & 44.0m - 47.0m & 47.0m & 33.0 - 35.0m\\

\hline
Representative LOD1 height & 46.0m & 47.0m & 34.0m  \\
\hline
Indicative ground elevation & 26.0m & 31.0m  & 24.0m \\
Indicative relative building height & 20.0m & 16.0m & 10.0m \\
\hline
\end{tabularx}
\label{table:heights}
\end{table}

To calculate the intersection height at which the LOS component of each signal would intersect the building, it was assumed that the signal travelled in a direct line with respected to the projected plane coordinate system. This neglects earth curvature, however for all plausible observations with at most 1km distance between an observer and a building, the effect is only 8cm. Likewise curvature of the signal due to atmospheric diffraction is less than 0.8 arcseconds, which equates to less than 1cm effect at 1 km \parencite{moller2019atmospheric}.

\subsection{Results}
\subsubsection{Signal Classifier}
GNSS mapping algorithms rely on a signal classifier to predict whether a signal was open (LOS or multipath with a LOS component) or closed (blocked, NLOS or multipath without a LOS component), and the performance of the signal classifier plays an important role in the quality of the produced map. It is important to investigate if bias and uncertainty in the signal classifier affects the produced map.  

This paper classifies individual signals by their \CN\ through implementing a 4PL probabilistic classifier, with the most likely classification being assigned. The form of classifier was chosen based on the results of \textcite{wang2015smartphone}. They proposed a quadratic spline form however we suggest that a 4PL has advantages of monotonicity and smoothness that make model fitting through optimisation more robust. It is worth reminding readers that this has no connection to the 4PL used in the map classifier, but happens to have a form that is a good fit to the observed data. More complex alternative classifiers are discussed in section \ref{gnss}, and use additional features available as raw data. Likewise, the \CN\ classifier could be extended to model ionospheric and tropospheric effects as well as taking into account specific device receiver gain and satellite transmission power. These complexities were not implemented in the paper, as the advantage of the proposed algorithm is that the generated map is relatively robust to the accuracy of the classifier.

\begin{figure}[!ht]
\centering
\includegraphics[width=0.5\linewidth]{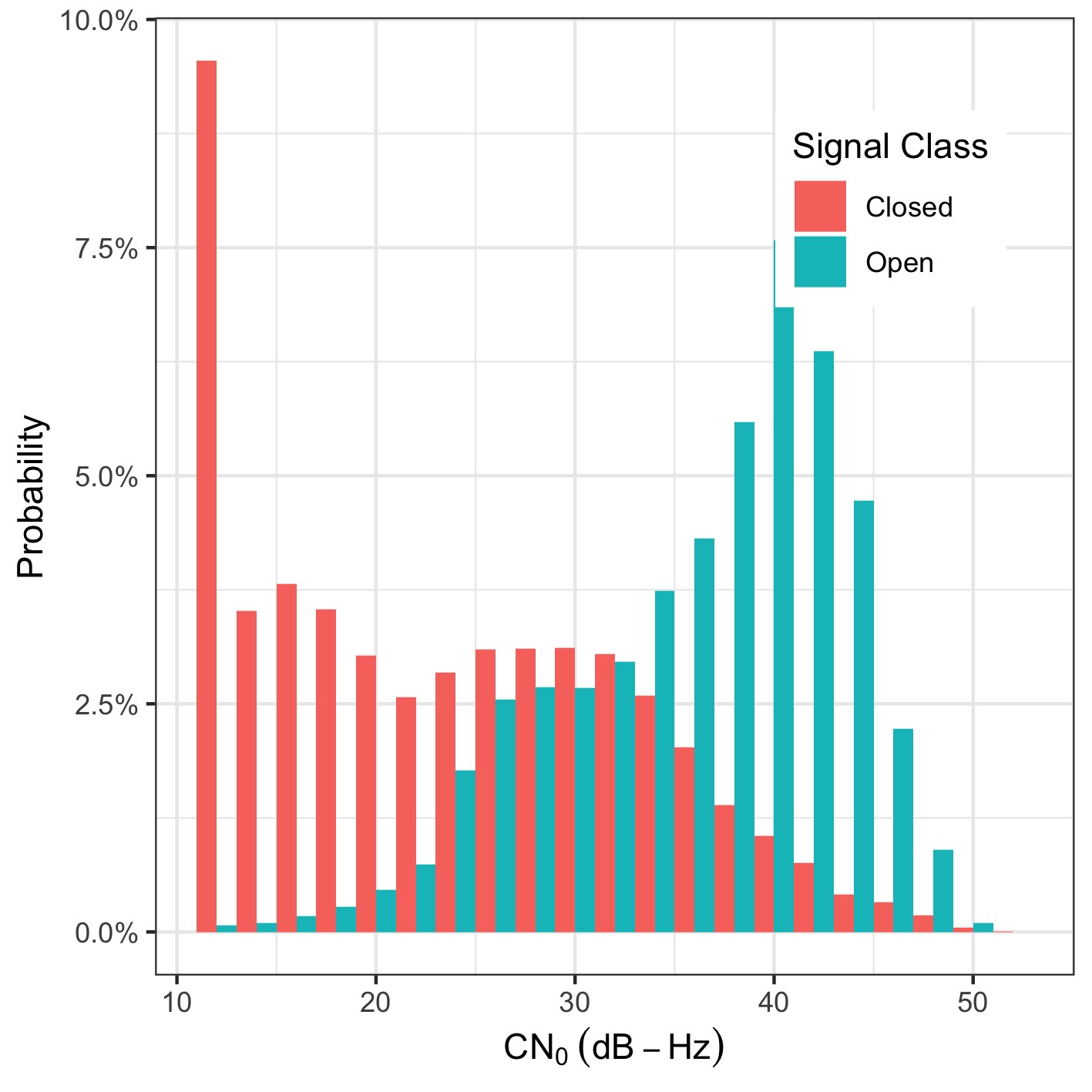}
\caption{Distribution of received signals}
\label{fig:ssdistribution}
\end{figure}

To assess the performance of the signal strength classifier, the ground truth building heights were used to label the dataset. The \CN\ distribution of open and closed signals is shown in figure \ref{fig:ssdistribution}. The two signal classes have distinct but overlapping distributions, demonstrating the difficulty of accurately predicting class. Predicting closed signals at low signal strengths can be more straightforward than open signals at high signal strengths, based on the overlap of the two distributions. Almost no signals were received with a \CN\ of less than 10 dB-Hz, which was an expected effect marking the cutoff below which the signal is too weak to be registered by the receiver. There is an anomalous peak of observations just above this level, which we are unable to explain with any certainty. 

\begin{figure}[h!]
\centering
\includegraphics[width=0.5\linewidth]{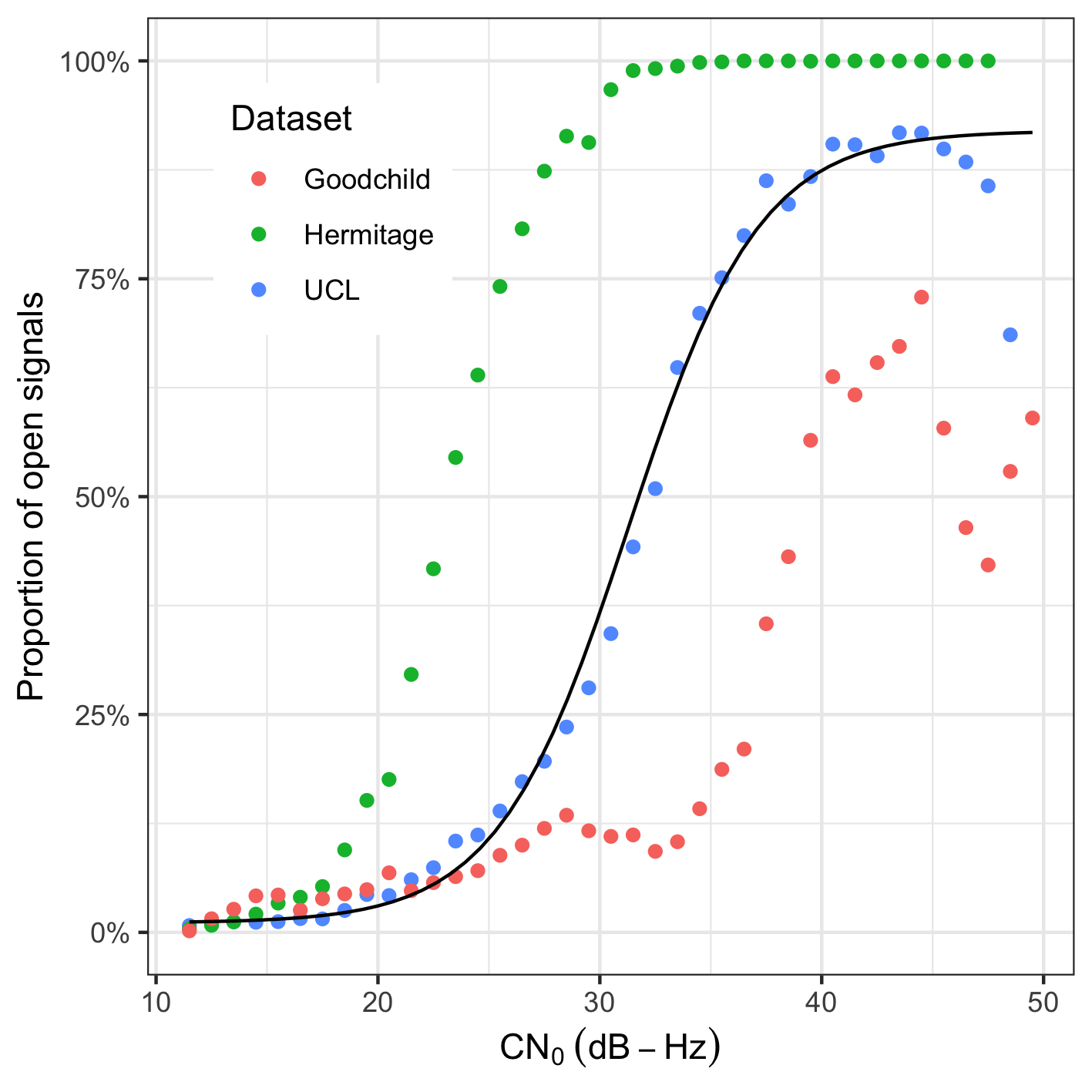}
\caption{Distribution of received signals}
\label{fig:ssfitted}
\end{figure}

A probabilistic 4PL classifier was trained on the dataset pertaining to the UCL building, and tested on the data from the other locations. Figure \ref{fig:ssfitted} illustrate how the varying observational settings affect the relationship between LOS probability and signal strength. which decrease the performance of trained signal classifiers. This leads to a stark divergence in performance of the signal strength classifier between training and test data.

\FloatBarrier

\begin{figure}[h!]
\centering
\includegraphics[width=0.8\linewidth]{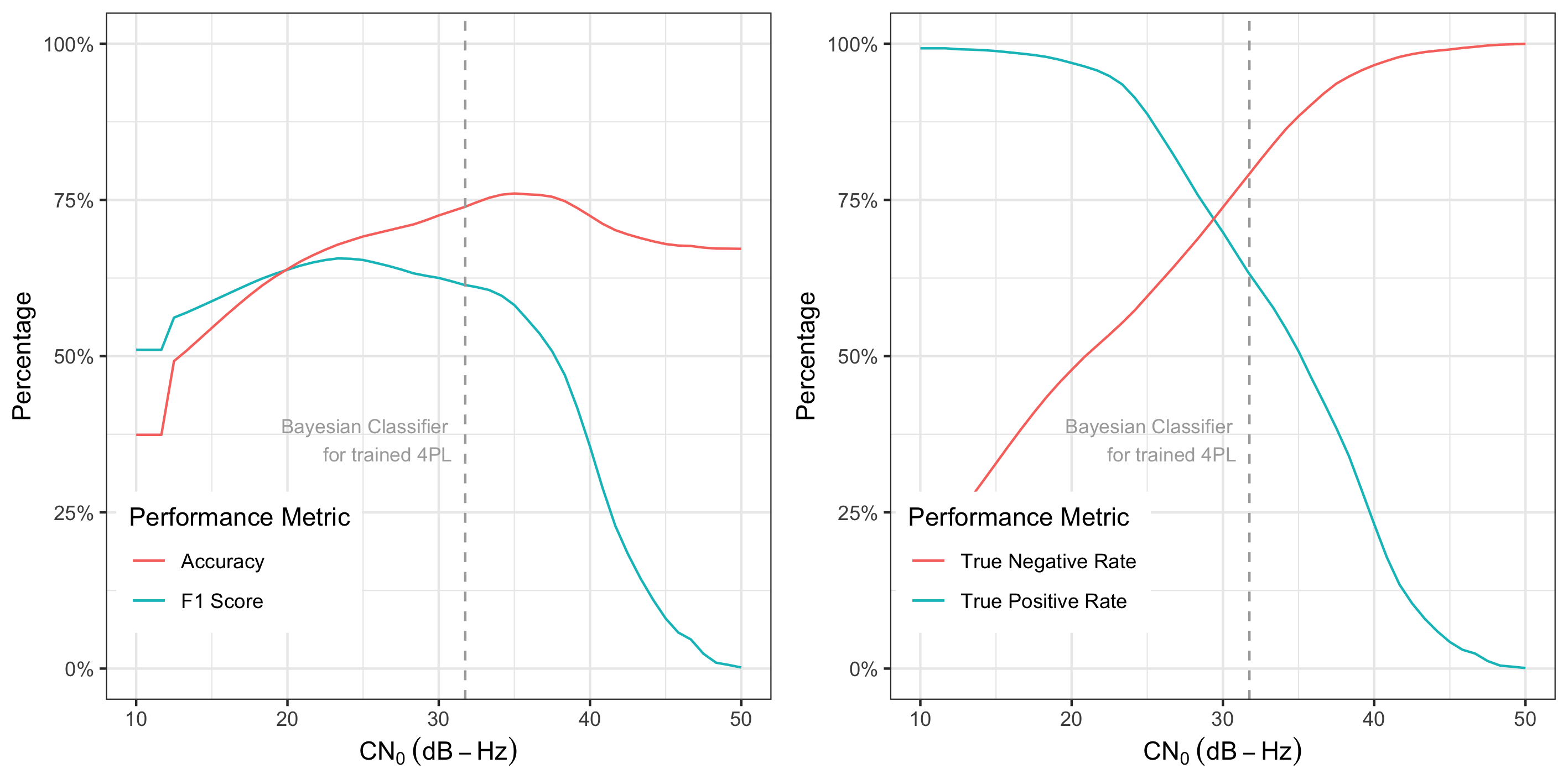}
\caption{Signal classifier performance with varying classification thresholds. Highlighted classification value represents the 50\% probability class threshold estimated by the trained 4PL}
\label{fig:ssperformance}
\end{figure}

The 4PL signal classifier fits the observed proportions for the training data relatively well, with a Mcfadden's $R^2$ of $70.7\%$ , but is wrongly parameterised for the test data, with a Mcfadden's $R^2$ of $18.9\%$. The performance of the 4PL across the dataset is shown in table \ref{table:ssclasses}, which cross-tabulates the true labels against the predictions obtained by the trained 4PL, using a 50\% probability threshold to label classes. Figure \ref{fig:ssperformance} illustrates the effect on accuracy of changing the classification threshold. The performance of a single classifier across the combined test locations can only be improved marginally by optimising parameters. As discussed in section \ref{gnss}, these parameters are device and location specific, making it difficult to generalise the classifiers. 

\begin{table}[h!]
\centering
\setlength\extrarowheight{6pt}
\small
\caption{Test data cross-tabulation by predicted (\CN) and actual (Height) class. Height n/a indicates the signal did not intersect the building.}
\begin{tabularx}{0.5\textwidth}{|
>{\hsize=0.9\hsize}X|
>{\hsize=0.9\hsize}X|
>{\hsize=1\hsize}X
>{\hsize=1\hsize}X
>{\hsize=1\hsize}X|
>{\hsize=1.2\hsize}X|
}
\hline
\multirow{2}{4em}{} & \multirow{2}{4em}{} & \multicolumn{3}{c}{Height Class} &\\
\cline{3-6}
 & & Open & Closed & n/a  & Total\\
\cline{3-6}
\multirow{2}{4em}{\CN\ Class} & Open & 40,683 & 27,474 & 95,062 & 163,219\\
& Closed & 23,750 & 104,339 & 294,780 & 422,869 \\
\cline{3-6}
& Total  & 64,433 & 131,813 & 389,842 & 586,088\\
\hline
\end{tabularx}
\label{table:ssclasses}
\end{table}

\FloatBarrier

\subsubsection{Map Algorithm}
To assess the map algorithm, it was applied to each location's dataset. To evaluate the impact of a misspecifed signal classifier, it was repeated using a range of parameters for the initial signal strength classifier. For comparison, the same signal classifier was used in the non-bootstrapped 4PL algorithm, the \textcite{weissman20132} hinge-loss method, and a Bayesian probabilistic method of maximising likelihood similar to \textcite{irish2014belief}, albeit applied to the entire dataset rather than to each voxel. The signal strength 4PL was initialised with the same initial parameters ($a$: 0.9, $b$: 0.2, $d$: 0.1) except for the inflection point, $c$, which ranged from $20$-$40$ dB-Hz in $1$ db-Hz increments. This generated a range of $50\%$ probability classification thresholds from $20$ to $40$ dB-Hz, which we believe is a plausible range: the thresholds for our datasets were between $23$ and $40$ dB-Hz respectively, and other works have ranged from $23$ dB-Hz \parencite{wang2015smartphone} to $37.5$ dB-Hz \parencite{yozevitch2016robust}. The bootstrapping algorithm was stopped at the lower of 10 iterations and fewer than 1\% of the labels having changed between two iterations of the map classifier.

\begin{figure}[h!]
\centering
\includegraphics[width=1\linewidth]{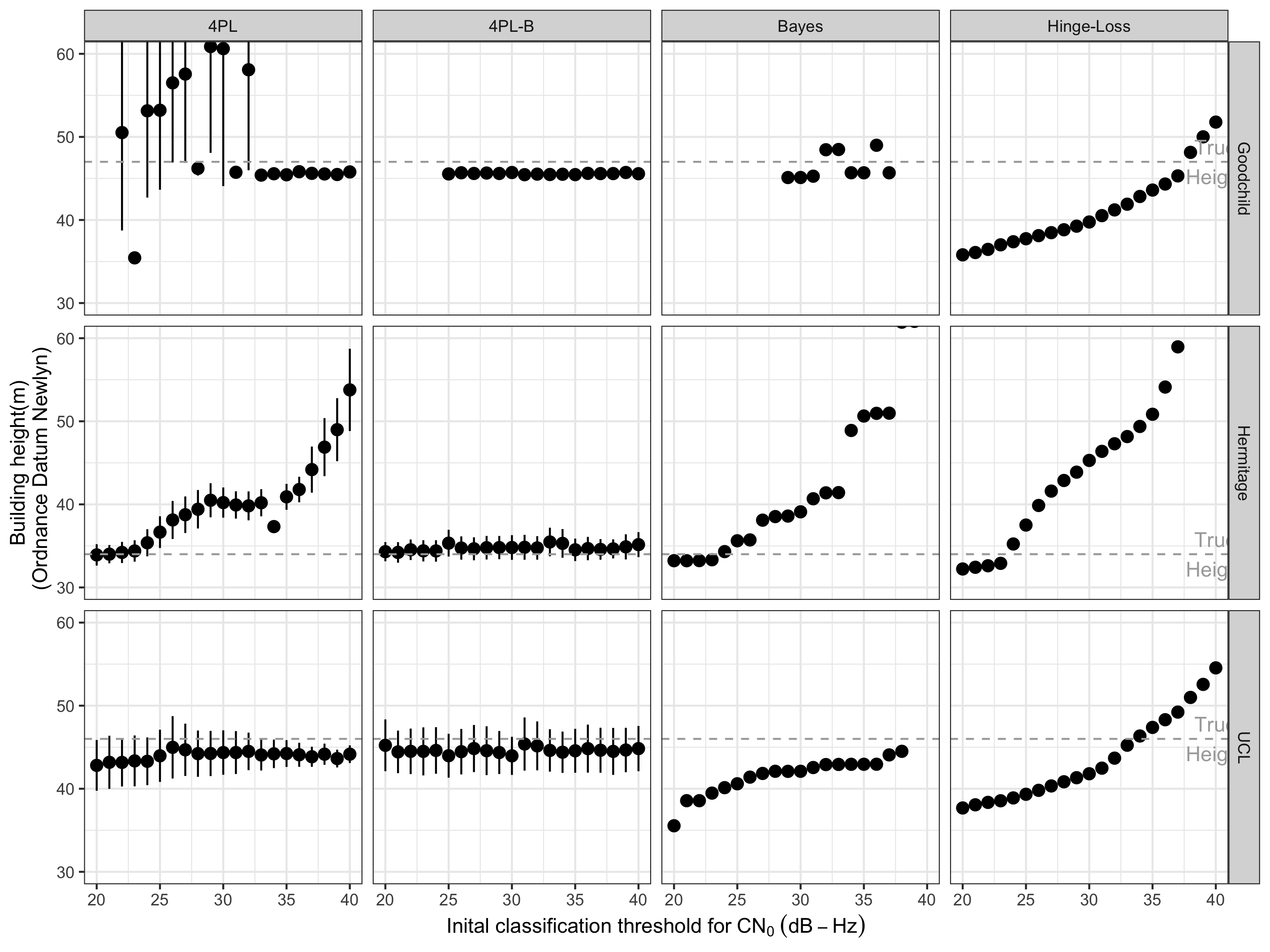}
\caption{Mapping algorithm results with varying signal strength classifiers}
\label{fig:algoperformance}
\end{figure}

Figure \ref{fig:algoperformance} shows the algorithms results when applied to the datasets with the manually inputted, more accurate, receiver locations. The 4PL-B algorithm converged to a solution at all 20 starting conditions at the UCL and Hermitage Road sites, and 15 out of 20 at the Goodchild Road sites. The other non-iterative algorithms always generated a solution (not always shown in the figure due to being outside the limits of the y-axis) however is not necessarily advantageous to return a low quality estimate. Table \ref{table:results} show the root mean square error for each algorithm and site across all 20 different initialisations. The 4PL-B algorithm is more accurate by an order of magnitude due to its consistency.

\begin{table}[h!]
\centering
\small
\caption{Root Mean Square Error of mapping algorithms}
\begin{tabularx}{0.75\textwidth}{|l|X|X|X|}
\hline
 & UCL & Goodchild Road & Hermitage Road\\
\hline\hline
Manually input locations &&&\\
4PL-B & 1.4m & 1.4m & 0.8m \\
4PL & 2.1m & 11.7m & 7.8m\\
Hinge-Loss & 5.6m & 7.4m & 20.6m\\
Bayes & 8.6m & 24.9m & 18.7m \\ 
\hline
GNSS position solutions &&&\\
4PL-B & 0.6m & - & 0.8m \\
4PL & 1.4m & 28.7m & 19.5m\\
Hinge-Loss & 5.6m & 10.1m & 9.9m\\
Bayes & 7.6m & 29.9m & 31.6m \\ 
\hline
\multicolumn{4}{p{0.7\textwidth}}{Calculated for each site across all twenty initialisations. 4PL-B results were excluded where the algorithm did not converge: 5 results for the Goodchild Road site using manual locations; all results at Goodchild Road using GNSS locations; and 4 at the Hermitage Road site using GNSS locations.}\\
\end{tabularx}
\label{table:results}
\end{table}
\FloatBarrier

Unlike the 4PL-B algorithm, which provides consistent results regardless of initialisation, results for the other methods depend on choice of initialisation. The Weissman hinge-loss algorithm increases superlinearly with initial classification threshold and nothing can be usefully inferred without a knowledge of the optimal classification threshold for the dataset. The Bayesian algorithm results increase monotonically, not smoothly but with stepped increases and relatively stable clustering around certain values. It requires less precise knowledge of the optimal classification threshold but the initial classification must still be relatively accurate otherwise the algorithm can generate wildly inaccurate results. The non-bootstrapped 4PL algorithm performance varies by location. In less challenging locations (UCL) it performs consistently, but in more challenging locations it requires a relatively accurate starting point, similar to the Bayesian approach.

The difficulty that the non-bootstrapped 4PL algorithm has in certain locations appears to be due to the overall balance of collected data. In the two more challenging locations, the environment and data collection protocol was such that most of the data was blocked (Goodchild Road) or open (Hermitage Road). The model is fit to the data as a maximum likelihood estimate with the simplifying assumption that signal strength is independent of height within a class. However there will always be weak patterns of height dependency and signal strength, and if most of the data is within the same class then the model fits weak patterns within the class, due to the weight of data, rather than the stronger pattern across classes which is only present in a small quantity of data. Preliminary work suggests that improvements can be obtained by filtering and rebalancing the data based on the distribution of intersection heights. Improvements to the 4PL algorithm could help broaden the conditions of convergence for the bootstrapped algorithm.

Table \ref{table:uncertainty} shows the 4PL-B algorithm measure of uncertainty around the building height based on the steepness of the 4PL curve, compared to the true height. The algorithm uncertainty is in accordance with the relative complexity of the building shape: Goodchild Road is a simple flat roof, Hermitage Road is a terrace with pitched roofs, and UCL has a complex roof shape. 

\begin{table}[h!]
\centering
\small
\caption{Uncertainty reported by 4PL-B algorithm}
\begin{tabularx}{0.75\textwidth}{|l|X|X|X|}
\hline
& UCL & Goodchild Road & Hermitage Road \\
\hline
Manually input locations & & & \\
min-max heights & 41.9m - 47.4m  & 45.4m - 45.8m & 33.4m - 36.2m \\
uncertainty range & 5.5m & 0.4m & 2.8m  \\
\hline
GNSS position solutions  & & & \\
min-max heights & 41.5m - 49.5m  & - & 30.7m - 38.8m \\
uncertainty range  & 8.0m & - & 8.1m  \\
\hline
Google Earth  & & & \\
min-max heights & 44.0m - 47.0m & 47.0m & 33.0 - 35.0m\\
uncertainty range &  3.0m & 0.0m & 2.0m\\

\hline
\multicolumn{4}{p{0.7\textwidth}}{Google Earth uncertainty for UCL height is that of the Library entrance hall. The entire building has a height which ranges between 26.0m - 58.0m \parencite{OSMastermap}.}\\
\end{tabularx}
\label{table:uncertainty}
\end{table}
\FloatBarrier

\begin{figure}[h!]
\centering
\includegraphics[width=0.8\linewidth]{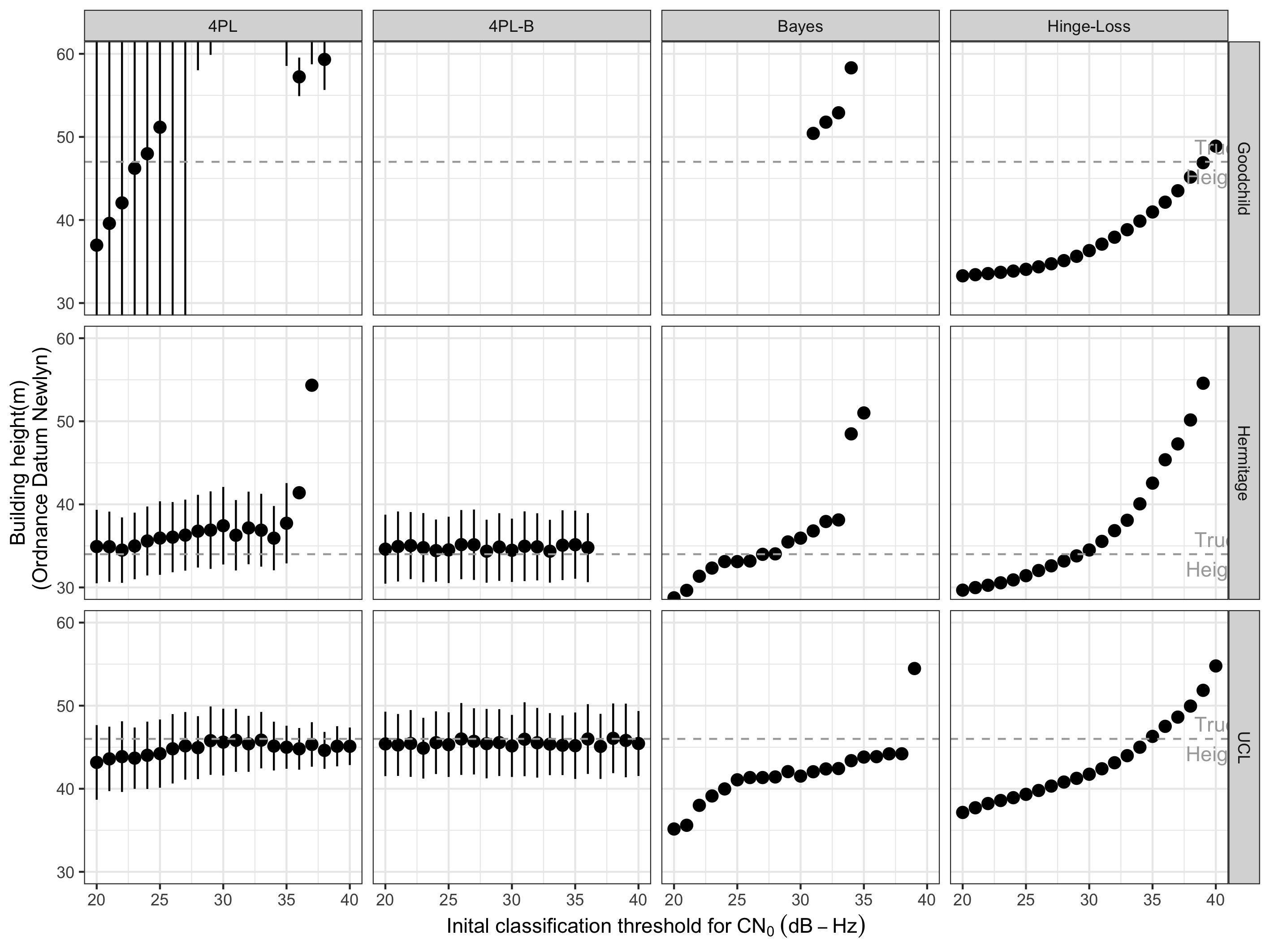}
\caption{Mapping algorithm results with GNSS position solutions}
\label{fig:algognss}
\end{figure}

The process was repeated on the data using the GNSS-provided latitude and longitudes rather than the manually input locations. This increases location errors but fully automates data-collection. We re-ran results using the GNSS-provided latitude and longitudes, and setting the altitude for all signals as 1m above ground level for the experimental location, using the OS Terrain 5 Digital Elevation Model \parencite{OSDTM}. By using digital elevation models, it is possible to avoid typically large vertical errors in GNSS positioning, which can create intersection height errors. The results are shown in Figure \ref{fig:algognss} and Table \ref{table:results}. The 4PL-B algorithm perform relatively accurately at the Hermitage and UCL sites, where satellites were visible at low and moderate elevations but failed completely for the Goodchild site, where only high elevation signals were visible. This is understandable because when only high elevation signals are visible, the GNSS position errors are typically larger due to geometric dilution of precision, and this is compounded because the high elevation also increases the effect of horizontal error on calculated intersection height. The other algorithms showed the same effects. The uncertainty reported by the 4PL-B increased with the GNSS solutions (table \ref{table:uncertainty}) highlighting that the algorithm is able to reflect the quality of the position data in its reported results.

\FloatBarrier

\section{Discussion} \label{discussion}
This section considers how the GNSS mapping algorithm could be scaled up to a VGI system on a city-wide scale. It discusses how a system could be implemented; how the algorithm could be extended to consider occlusion; whether the approach extends to a higher LOD; and the role of location errors.

The key advantage of a GNSS mapping approach is that participation in the VGI project is straightforward and almost entirely passive. The necessary data is already continually collected by GNSS-enabled smartphones to facilitate positioning and the use of location-based services. Since version 7.0 introduced in 2016, the Android operating system allows access to “raw” GNSS data through the android.location  Application Programming Interface (API) \parencite{ESAraw}, i.e. any installed mobile applications (app) with user permissions can use the GNSS raw data. The API usually provides signal observations at a 1Hz frequency, and consequently a single smartphone can generate above 100,000 signal observations an hour, including signals that were expected to be received but in practice were blocked due to obstacles between the satellites and the receivers such as buildings. 

To implement a large-scale VGI system primarily requires the creation of Android mobile app, which is installed by a participant, and for it to run as needed in the background, with the option of the user manually inputting locations. The app can then periodically upload data to a central server for processing. GNSS data provided through the API includes all the information required by the various signal classifiers described in Section \ref{gnss}: the GNSS navigation message, from which the satellite and its azimuth and elevation can be identified, along with typical GNSS observables such as pseudorange, carrier phase, precise timing and doppler shift. The carrier-to-noise ratio is also available. Further study is planned to develop a VGI mobile application and address pertinent questions e.g. privacy, incentivisation and participation.

The presented algorithm considers signals being blocked by a single building. On a larger scale, additional occlusion may occur. This could be due to a taller building further away from the receiver in the same direction, or unmapped objects (e.g. buildings due to a outdated basemap; or trees and other street furniture) between the receiver and the building of interest. This would present itself as greater uncertainty and error in the height estimate due to the occlusion effects. 

Spatiotemporal diversity of data collection should mitigate many of these errors, as the occlusion will vary based on the receiver position and signal azimuth and elevation. This suggests a more granular extension of the algorithm which groups signals by their direct path. This would lend itself to a higher LOD map, as the estimated height can vary across the building, and also a visibility model where the height estimates from different groups of signals can be compared, and outliers would suggest an occlusion which can then be taken into account by amending the basemap or applying the algorithm to a filtered dataset. 

The algorithm itself as a photo-consistency measure can be improved: the four-parameter logistic is fit to the data as a maximum likelihood estimate. If most of the data is within the same class (either because the building is tall and close-by, or small and measured from a distance) then the model can fit weak patterns within the class, due to the weight of data, rather than the stronger pattern of interest which is only present in a small quantity of data. An extension of the algorithm could attempt to reweight the data as the height estimate develops. 

Location error is an important component of the algorithm performance. In urban canyons (narrow streets with tall buildings on either size) it can be expected that the algorithm performance degrades. This is because LOS to GNSS satellites is only possible at a high elevation. This leads to poor cross-street position accuracy \parencite{groves2012intelligent} and the effect of horizontal error on calculated intersection height is magnified. Further study is planned to extend the current algorithm to improve its accuracy in this context: modelling location error and signal diffraction, and using a more complex signal classifier.

\section{Conclusion}
This paper proposed a novel algorithm that automatically generates 3D LOD 1 maps by processing GNSS data. The technique works on unlabelled signal observations by co-training a signal classifier and having a robust design to any potential bias in the signal classifier. The implementation of the algorithm shows that it is unaffected by bias in the initial data labelling and the results are comparable to existing benchmarks for 3D maps \parencite{groger2012ogc}, although larger scale data collection is required to provide further experimental verification. By providing a low-cost alternative, this work could be very helpful for the large scale production of 3D maps, particularly for regions where the cost of existing technologies is prohibitive. These maps have many useful applications in location-based services, autonomous vehicle and drone navigation, and urban planning. 

\section*{Acknowledgements}
We thank UK Research and Innovation for their support. The work presented in this paper has been funded by the UK Research and Innovation (UKRI) Future Leaders Fellowship MR/S01795X/2 

\printbibliography

\end{document}